\documentclass[letterpaper, 10 pt, conference]{ieeeconf}  

\IEEEoverridecommandlockouts                              

\usepackage{times}
\usepackage[shortcuts]{extdash} 

\usepackage{multicol}
\usepackage[bookmarks=true]{hyperref}

\usepackage[utf8]{inputenc} 
\usepackage[T1]{fontenc}    
\usepackage{hyperref}       
\usepackage{url}            
\usepackage{booktabs}       
\usepackage{amsfonts}       
\usepackage{nicefrac}  
\usepackage{xcolor}
\usepackage{microtype}      
\usepackage{lipsum}
\usepackage{graphicx}
\usepackage{subfigure}
\usepackage{multicol}  
\usepackage{multirow}
\usepackage{stfloats}
\usepackage{textcomp}

\usepackage{amsmath,amssymb,amsfonts}
\usepackage{algorithmic}
\usepackage{textcomp}
\usepackage{xcolor}
\usepackage{algorithm}
\usepackage{multicol}  
\usepackage{multirow} 
\usepackage{subfigure}
\usepackage{dashrule}

\usepackage{listings}
\usepackage{listings}
\usepackage{xcolor}
\usepackage{pdfpages}

\definecolor{codegreen}{rgb}{0,0.6,0}
\definecolor{codegray}{rgb}{0.5,0.5,0.5}
\definecolor{codepurple}{rgb}{0.58,0,0.82}
\definecolor{backcolour}{rgb}{0.95,0.95,0.92}

\lstdefinestyle{python}{
    backgroundcolor=\color{backcolour},   
    commentstyle=\color{codegreen},
    keywordstyle=\color{magenta},
    numberstyle=\tiny\color{codegray},
    stringstyle=\color{codepurple},
    basicstyle=\ttfamily\footnotesize,
    breakatwhitespace=false,         
    breaklines=true,                 
    captionpos=b,                    
    keepspaces=true,                 
    numbers=left,                    
    numbersep=5pt,                  
    showspaces=false,                
    showstringspaces=false,
    showtabs=false,                  
    tabsize=2
}

\begin{document}

\title{DINOv3-Diffusion Policy: Self-Supervised Large Visual Model for Visuomotor Diffusion Policy Learning}


\author{ThankGod Egbe, Peng Wang, Zhihao Guo, Zidong Chen
\thanks{Thankgod Egbe, Zhihao Guo, and Peng Wang are with the Department of Computing and Mathematics, Manchester Metropolitan  University, Manchester, M15 6BH, UK. Zidong Chen is with the Department of Computing, Imperial College London, London, SW7 2AZ, UK. 
{\tt\small Contact: \{thankgod.egbe, zhihao.guo\}@stu.mmu.ac.uk}, zidong.chen25@imperial.ac.uk, p.wang@mmu.ac.uk}%
}

\maketitle

\begin{abstract}

This paper evaluates DINOv3, a recent large-scale self-supervised vision backbone, for visuomotor diffusion policy learning in robotic manipulation. We investigate whether a purely self-supervised encoder can match or surpass conventional supervised ImageNet-pretrained backbones (e.g., ResNet-18) under three regimes: training from scratch, frozen, and finetuned. Across four benchmark tasks (Push-T, Lift, Can, Square) using a unified FiLM-conditioned diffusion policy, we find that (i) finetuned DINOv3 matches or exceeds ResNet-18 on several tasks, (ii) frozen DINOv3 remains competitive, indicating strong transferable priors, and (iii) self-supervised features improve sample efficiency and robustness. These results support self-supervised large visual models as effective, generalizable perceptual front-ends for action diffusion policies, motivating further exploration of scalable label-free pretraining in robotic manipulation. Compared to using ResNet18 as a backbone, our approach with DINOv3 achieves up to a 10\% absolute increase in test-time success rates on challenging tasks such as Can, and on-the-par performance in tasks like Lift, PushT, and Square.

\end{abstract}

\IEEEpeerreviewmaketitle

\section{Introduction}

\textbf{RQ}: \textit{How does incorporating a novel DINOv3 self-supervised vision backbone affect policy performance and generalisation in visuomotor policy learning via action diffusion, compared to standard supervised vision backbones? }

For decades, advances in machine learning research, focused on deep learning, have revolutionised fields such as computer vision and natural language processing \cite{dosovitskiy_image_2021},\cite{vaswani_attention_2023}. Whereas traditional machine learning methods rely on training models using pre-constructed datasets for pattern recognition and prediction, embodied learning emphasises knowledge acquisition through physical interactions and practical experiences \cite{roy_machine_2021}. Advanced perception capabilities and precise policy generation are fundamental components of learning from demonstration, or imitation learning, a crucial pillar in object-centric embodied learning for robotic manipulation \cite{zheng_survey_2025}. To address key aspects of perception and policy learning in embodied robotic systems, recent research has increasingly focused on cutting-edge technologies such as Large Language Models (LLMs) \cite{zhao_survey_2025, wang2024llm, xiao2023llm}, Diffusion Models \cite{ho_denoising_2020, wang2024robot}, 3D Gaussian Splatting \cite{kerbl_3d_2023}, Neural Radiance Fields (NeRFs) \cite{mildenhall_nerf_2020}, and others. These innovations collectively push the boundaries of how robots perceive, represent, and act within complex environments by enhancing semantic understanding, spatial reasoning, and precise control capabilities.

Visuomotor policy learning via action diffusion introduces a novel class of robotic manipulation policies derived from demonstrations, grounded in a conditional denoising diffusion process~\cite{chiDiffusionPolicyVisuomotor2024}. This approach is inspired by the success of Denoising Diffusion Probabilistic Models (DDPMs) in generative image modelling \cite{ho_denoising_2020}, adapting the diffusion framework to model complex, multimodal robot action distributions conditioned on visual inputs. The diffusion policy work has made groundbreaking contributions by developing a generative action policy that supports visual conditioning, enabling real-time action inference and closed-loop action sequences. This design ensures temporal action consistency, effectively balancing long-horizon planning with responsiveness during manipulation tasks. 

To perform object-centric robotic manipulation via visuomotor policy learning, the robot must first accurately perceive the target object and its surrounding environment. Visual encoders map the raw image sequence from as many camera views into a latent embedding, comprising the latent observation embedding for diffusion policy learning. This structured representation, often combined with language-based modalities, supports recent advances in visuomotor policy learning by providing the essential perceptual and language grounding needed for robust control and informed decision-making in complex, dynamic environments \cite{zheng_survey_2025}.

Vision-based robot policy learning typically employs one of three encoder usage regimes: (i) training from scratch with randomly initialized weights, (ii) freezing a pretrained encoder as a fixed feature extractor, and (iii) end-to-end fine-tuning of pretrained weights~\cite{chiDiffusionPolicyVisuomotor2024}. Prior work has mainly adopted two backbone families: lightweight convolutional networks (e.g., ResNet-18/34~\cite{he_deep_2015}) and Vision Transformers (e.g., ViT-B/16~\cite{dosovitskiy_image_2021}). CNNs provide strong inductive biases for locality and translation, while ViTs offer flexible global receptive fields and improved multi-object spatial reasoning. When paired with appropriate pretraining, both classes yield dense representations that support stable conditioning of diffusion-based visuomotor policies~\cite{chiDiffusionPolicyVisuomotor2024, chi_universal_2024}.

Supervised large-scale pretraining on datasets such as ImageNet-1K and -21K~\cite{krizhevsky_imagenet_2017, ridnik_imagenet-21k_2021} has long underpinned robotic perception, with ResNet variants commonly fine-tuned or partially adapted for manipulation tasks. Domain-aligned representation learning has further advanced through robotics-relevant pretraining (e.g., R3M leveraging egocentric video), and multimodal vision–language encoders such as CLIP and SigLip, which have improved generalisation in recent policy frameworks~\cite{chi_universal_2024}. 

In parallel, self-supervised vision backbones (DINO to DINOv2) have shown strong transfer without manual labels, and have been selectively integrated as auxiliary components in broader visuomotor stacks (e.g., OpenVLA using DINOv2 or SigLip for spatial grounding~\cite{kim_openvla_2024}). Recently, DINOv3~\cite{simeoni2025dinov3} has emerged as a state-of-the-art self-supervised vision model, trained on an unprecedentedly large and diverse dataset (1.7B images) using a scalable distillation approach. DINOv3 demonstrates superior performance across various vision benchmarks, including image classification, segmentation, and retrieval, outperforming prior self-supervised models and even some supervised counterparts. Its robust feature representations and scalability make it a promising candidate for enhancing visuomotor policy learning in robotics. To the best of our knowledge, DINOv3 has not yet been systematically evaluated as a vision backbone within action diffusion–based visuomotor policy learning frameworks, and this work aims to fill that gap.


\begin{figure*}[t!]
    \centering
    \includegraphics[width=0.90\linewidth]{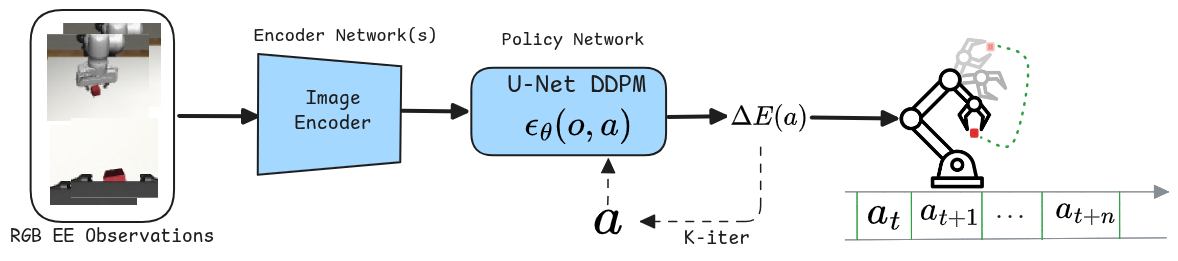}
    \caption{General diffusion policy training pipeline using DINOv3 as vision embedding model for action diffusion. RGB end-effector observations are processed through a DINOv3-based image encoder to extract visual features, which condition a U-Net DDPM policy network $\epsilon_\theta(o,a)$. The policy network iteratively denoises action sequences through $k$ diffusion steps, predicting noise $\Delta E(a)$ to generate a sequence of actions $a_t, a_{t+1}, \ldots, a_{t+n}$ for robotic control.}
    \label{fig:pipeline}
\end{figure*}

\begin{figure*}[t!]
    \centering
    \includegraphics[width=0.90\linewidth]{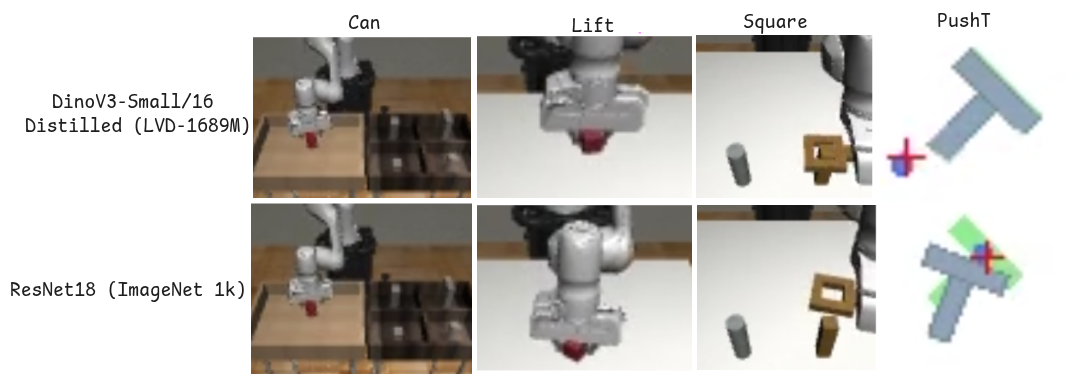}
    \caption{Simulation rollout from 10,000 optimisation steps each showing the DINOv3 model consistently faster and more concise than ResNet18 variant}
    \label{fig:task}
\end{figure*}

\section{Approaches}
In line with the research question, we designed simulation experiments using four tasks, namely Push-T, Can, Lift, and Square, all selected from two established simulation benchmarks \cite{florence_implicit_2021, mandlekar_what_2021} and adapted from prior work \cite{chiDiffusionPolicyVisuomotor2024}. These experiments evaluate the performance of both ResNet-18 and ViT-small/16 distilled encoder architectures, pretrained on ImageNet-1K and Dinov3 LVD1689M datasets, respectively, as vision backbones showing in Fig.~\ref{fig:pipeline}. This setup enables a comparative analysis of conventional supervised pretrained encoders against SOTA self-supervised models within visuomotor policy learning frameworks. Each vision encoder model architecture based on a separate observation input is evaluated under three training strategies: end-to-end training with randomly initialised weights, using frozen pretrained weights, and fine-tuning pretrained encoders. These strategies are implemented on a CNN-based U-Net architecture \cite{ronneberger_u-net_2015} with FiLM (Feature-wise Linear Modulation \cite{perez_film_2018}) observation embedding conditioning, known for its more stable training dynamics compared to multi-head cross-attention transformer-based \cite{vaswani_attention_2023} diffusion policy models \cite{chiDiffusionPolicyVisuomotor2024}. 

Each experiment is trained for a minimum of 100 epochs to achieve significant training time savings, with checkpoints saved every 20 epochs. Training beyond 500 epochs showed clear signs of overfitting on the validation loss, with no meaningful improvements in test success rates even after extending training up to 3,000 epochs. All experiments are conducted on a Nvidia RTX A6,000 Ada Gen GPU with 48GB memory.

\section{Results and Discussions}

\begin{table*}[ht!]
  \centering
  \caption[Consolidated Experiment Results]{Comparison of experiment results across tasks (Train and Test). All Robomimic \cite{mandlekar_what_2021} is based on ph dataset no mh dataset was used for this experiment. Each experiment was trained for 100 epochs across all task using both ResNet18 and ViT small (DINOv3) model architecture}
  \label{table:all_experiments}
  \begin{tabular}{c|c|c|c|c|c|c|c}
    \toprule
    \multirow{2}{*}{Task} & \multirow{2}{*}{Architecture} & \multirow{2}{*}{Split} & \multirow{2}{*}{No Pretrained} & \multicolumn{4}{c}{Pretrained} \\
                          &                                &                        &                                & Dataset & Parameters & Frozen & Finetuning \\
    \midrule
    \multirow{4}{*}{PushT}
        & \multirow{2}{*}{ResNet18} 
            & Train & \textbf{0.81} & Imagenet-1k & 11.7M & \textbf{0.65} & 0.79 \\
        &   & Test  & \textbf{0.78} & Imagenet-1k & 11.7M & \textbf{0.53} & 0.89 \\
        & \multirow{2}{*}{ViT-Small/16}
            & Train & 0.72 & LVD-1689M & 21M & 0.53 & \textbf{0.85} \\
        &   & Test  & 0.67 & LVD-1689M & 21M & 0.39 & \textbf{0.84} \\
    \midrule
    \multirow{4}{*}{Lift}
        & \multirow{2}{*}{ResNet18} 
            & Train & \textbf{1.00} & Imagenet-1k & 11.7M & \textbf{1.00} & \textbf{1.00} \\
        &   & Test  & \textbf{1.00} & Imagenet-1k & 11.7M & \textbf{1.00} & \textbf{1.00} \\
        & \multirow{2}{*}{ViT-Small/16}
            & Train & \textbf{1.00} & LVD-1689M & 21M & \textbf{1.00} & \textbf{1.00} \\
        &   & Test  & \textbf{1.00} & LVD-1689M & 21M & \textbf{1.00} & \textbf{1.00} \\
    \midrule
    \multirow{4}{*}{Can}
        & \multirow{2}{*}{ResNet18} 
            & Train & \textbf{1.00} & Imagenet-1k & 11.7M & \textbf{0.66} & 0.83 \\
        &   & Test  & \textbf{0.90} & Imagenet-1k & 11.7M & \textbf{0.70} & 0.80 \\
        & \multirow{2}{*}{ViT-Small/16}
            & Train & 0.33 & LVD-1689M & 21M & 0.50 & \textbf{1.00} \\
        &   & Test  & 0.50 & LVD-1689M & 21M & 0.50 & \textbf{0.90} \\
    \midrule
    \multirow{4}{*}{Square}
        & \multirow{2}{*}{ResNet18} 
            & Train & \textbf{0.83} & Imagenet-1k & 11.7M & 0.33 & \textbf{1.00} \\
        &   & Test  & \textbf{1.00} & Imagenet-1k & 11.7M & 0.10 & \textbf{1.00} \\
        & \multirow{2}{*}{ViT-Small/16}
            & Train & \textbf{0.83} & LVD-1689M & 21M & \textbf{0.33} & 0.83 \\
        &   & Test  & 0.50 & LVD-1689M & 21M & \textbf{0.60} & 0.90 \\
    \bottomrule
  \end{tabular}
\end{table*}

The results of our experiments are highlighted in Table \ref{table:all_experiments} and Figure \ref{fig:task} presents the results on both vision embedding models across all tasks. The comparative performance of ResNet-18 and DINOv3-small/16 vision backbones across four robotic manipulation tasks under different training strategies. For the PushT task, finetuning DINOv3 achieved the highest test success rate (0.84), surpassing both frozen and fine-tuned ResNet-18 models, demonstrating its superior adaptability in dynamic pushing scenarios. In the Lift task, both architectures performed exceptionally well across all strategies, reaching near-perfect success rates, indicating short-horizon task simplicity and possibly less sensitivity to the choice of vision backbone. For the Can task, DINOv3 finetuning notably outperformed ResNet-18, especially on the test set, where it reached a 0.90 success rate versus ResNet’s 0.80, underscoring DINOv3’s enhanced capability in precise object manipulation. Finally, on the Square task, finetuned ResNet-18 achieved perfect test performance, but DINOv3’s frozen and finetuned variants also showed strong improvements compared to no-pretraining conditions, suggesting robust spatial reasoning with DINOv3 embeddings. Overall, the results emphasize the strength of DINOv3, especially when fine-tuned, in providing rich visual features that translate to improved task execution across diverse manipulation scenarios.

\section{Conclusion} 
\label{sec:conclusion}
This paper presented a systematic evaluation of DINOv3, a SOTA self-supervised vision backbone, for visuomotor diffusion policy learning in robotic manipulation. Our experiments show that DINOv3, pretrained on a large and diverse unlabeled dataset, matches or surpasses the performance of conventional supervised ResNet-18 models pretrained on ImageNet-1K across a range of manipulation tasks. These findings demonstrate that self-supervised visual encoders like DINOv3 can provide robust, transferable representations that support efficient and effective visuomotor policy learning. To our knowledge, this is the first study to assess DINOv3 as a standalone vision backbone within action diffusion-based visuomotor policy frameworks, highlighting its promise for advancing label-free pretraining in robotics. Future work should explore the benefits of DINOv3 and similar models on more complex, long-horizon tasks~\cite{chiDiffusionPolicyVisuomotor2024}.

\section*{APPENDIX}

\subsection{Training configuration for finetuning DINOv3 small/16 distilled LVD1689M}\label{subsec:promptturtle}

\noindent
\begin{lstlisting}
task_name: lift
exp_name: default

horizon: 16
n_obs_steps: 2
n_action_steps: 8
n_latency_steps: 0
dataset_obs_steps: ${n_obs_steps}
past_action_visible: False
keypoint_visible_rate: 1.0
obs_as_global_cond: True
name: train_diffusion_unet_hybrid
shape_meta:
  action:
    shape: [10]
  obs:
    agentview_image:
      shape: [3, 84, 84]
      type: rgb
    robot0_eef_pos:
      shape: [3]  
    robot0_eef_quat:
      shape: [4]
    robot0_eye_in_hand_image:
      shape: [3, 84, 84]
      type: rgb
    robot0_gripper_qpos:
      shape: [2]
task:
  abs_action: ${abs_action}
  dataset:
    _target_: diffusion_policy.dataset.robomimic_replay_image_dataset.RobomimicReplayImageDataset
    abs_action: ${abs_action}
    dataset_path: ${dataset_path}
    horizon: 16
    n_obs_steps: 2
    pad_after: 7
    pad_before: 1
    rotation_rep: rotation_6d
    seed: 42
    shape_meta: ${shape_meta}
    use_cache: false
    val_ratio: 0.01
  dataset_path: ${dataset_path}
  #dataset_type: ph
  env_runner:
    _target_: diffusion_policy.env_runner.robomimic_image_runner.RobomimicImageRunner
    abs_action: ${abs_action}
    crf: 22
    dataset_path: ${dataset_path}
    fps: 10
    max_steps: 400
    n_action_steps: 8
    n_envs: null
    n_obs_steps: 2
    n_test: 10
    n_train: 6
    n_train_vis: 2
    n_test_vis: 2
    past_action: false
    render_obs_key: agentview_image
    shape_meta: ${shape_meta}
    test_start_seed: 100000
    tqdm_interval_sec: 1.0
    train_start_idx: 0
  name: lift_image
  shape_meta: ${shape_meta}
  task_name: ${task_name}
ema:
  _target_: diffusion_policy.model.diffusion.ema_model.EMAModel
  update_after_step: 0
  inv_gamma: 1.0
  power: 0.75
  min_value: 0.0
  max_value: 0.9999
dataloader:
  batch_size: 16 # try 64
  num_workers: 4 #try 8/12
  persistent_workers: false
  pin_memory: true
  shuffle: true
val_dataloader:
  batch_size: 16
  num_workers: 4 #try 8/12
  persistent_workers: false
  pin_memory: true
  shuffle: false
policy:
  _target_: diffusion_policy.policy.diffusion_unet_image_policy.DiffusionUnetImagePolicy
  cond_predict_scale: true
  diffusion_step_embed_dim: 64
  down_dims: [128, 256, 512]
  #eval_fixed_crop: true
  horizon: ${horizon}
  kernel_size: 5
  n_action_steps: 8
  n_groups: 8
  n_obs_steps: 2
  noise_scheduler:
    _target_: diffusers.schedulers.scheduling_ddpm.DDPMScheduler
    beta_end: 0.02
    beta_schedule: squaredcos_cap_v2
    beta_start: 0.0001
    clip_sample: true
    num_train_timesteps: 100
    prediction_type: epsilon
    variance_type: fixed_small
  obs_encoder:
    _target_: diffusion_policy.model.vision.multi_image_obs_encoder.MultiImageObsEncoder
    shape_meta: ${shape_meta}
    rgb_model:
      _target_: diffusion_policy.model.vision.model_getter.get_rgb_model
      name: "facebook/dinov3-vits16-pretrain-lvd1689m"
      weights: null
      pretrained: true
    resize_shape: null
    crop_shape: [76, 76]
    # constant center crop
    random_crop: True
    use_group_norm: True
    share_rgb_model: False
    imagenet_norm: True
  num_inference_steps: 100
  obs_as_global_cond: ${obs_as_global_cond}
  #obs_encoder_group_norm: true
  shape_meta: ${shape_meta}
training:
  device: "cuda:0"
  seed: 42
  debug: True
  resume: True
  # optimization
  lr_scheduler: cosine
  lr_warmup_steps: 500
  
  gradient_accumulate_every: 2
  # EMA destroys performance when used with BatchNorm
  # replace BatchNorm with GroupNorm.
  use_ema: True
  freeze_encoder: False #replace to True
  num_epochs: 100
  rollout_every: 20
  checkpoint_every: 20
  val_every: 1
  sample_every: 5
  # steps per epoch
  max_train_steps: null
  max_val_steps: null
  # misc
  tqdm_interval_sec: 1.0
optimizer:
  _target_: torch.optim.AdamW
  lr: 1.0e-4
  betas: [0.95, 0.999]
  eps: 1.0e-8
  weight_decay: 1.0e-6
\end{lstlisting}




\bibliographystyle{IEEEtran}
\bibliography{references}

\begin{thebibliography}{10}
\providecommand{\url}[1]{#1}
\csname url@samestyle\endcsname
\providecommand{\newblock}{\relax}
\providecommand{\bibinfo}[2]{#2}
\providecommand{\BIBentrySTDinterwordspacing}{\spaceskip=0pt\relax}
\providecommand{\BIBentryALTinterwordstretchfactor}{4}
\providecommand{\BIBentryALTinterwordspacing}{\spaceskip=\fontdimen2\font plus
\BIBentryALTinterwordstretchfactor\fontdimen3\font minus \fontdimen4\font\relax}
\providecommand{\BIBforeignlanguage}[2]{{%
\expandafter\ifx\csname l@#1\endcsname\relax
\typeout{** WARNING: IEEEtran.bst: No hyphenation pattern has been}%
\typeout{** loaded for the language `#1'. Using the pattern for}%
\typeout{** the default language instead.}%
\else
\language=\csname l@#1\endcsname
\fi
#2}}
\providecommand{\BIBdecl}{\relax}
\BIBdecl

\bibitem{dosovitskiy_image_2021}
\BIBentryALTinterwordspacing
A.~Dosovitskiy, L.~Beyer, A.~Kolesnikov, D.~Weissenborn, X.~Zhai, T.~Unterthiner, M.~Dehghani, M.~Minderer, G.~Heigold, S.~Gelly, J.~Uszkoreit, and N.~Houlsby, ``\BIBforeignlanguage{en}{An {Image} is {Worth} 16x16 {Words}: {Transformers} for {Image} {Recognition} at {Scale}},'' Jun. 2021, arXiv:2010.11929 [cs]. [Online]. Available: \url{http://arxiv.org/abs/2010.11929}
\BIBentrySTDinterwordspacing

\bibitem{vaswani_attention_2023}
\BIBentryALTinterwordspacing
A.~Vaswani, N.~Shazeer, N.~Parmar, J.~Uszkoreit, L.~Jones, A.~N. Gomez, L.~Kaiser, and I.~Polosukhin, ``Attention {Is} {All} {You} {Need},'' Aug. 2023, arXiv:1706.03762 [cs]. [Online]. Available: \url{http://arxiv.org/abs/1706.03762}
\BIBentrySTDinterwordspacing

\bibitem{roy_machine_2021}
\BIBentryALTinterwordspacing
N.~Roy, I.~Posner, T.~Barfoot, P.~Beaudoin, Y.~Bengio, J.~Bohg, O.~Brock, I.~Depatie, D.~Fox, D.~Koditschek, T.~Lozano-Perez, V.~Mansinghka, C.~Pal, B.~Richards, D.~Sadigh, S.~Schaal, G.~Sukhatme, D.~Therien, M.~Toussaint, and M.~V.~d. Panne, ``From {Machine} {Learning} to {Robotics}: {Challenges} and {Opportunities} for {Embodied} {Intelligence},'' Oct. 2021, arXiv:2110.15245 [cs]. [Online]. Available: \url{http://arxiv.org/abs/2110.15245}
\BIBentrySTDinterwordspacing

\bibitem{zheng_survey_2025}
\BIBentryALTinterwordspacing
Y.~Zheng, L.~Yao, Y.~Su, Y.~Zhang, Y.~Wang, S.~Zhao, Y.~Zhang, and L.-P. Chau, ``\BIBforeignlanguage{en}{A {Survey} of {Embodied} {Learning} for {Object}-{Centric} {Robotic} {Manipulation}},'' \emph{\BIBforeignlanguage{en}{Machine Intelligence Research}}, vol.~22, no.~4, pp. 588--626, Aug. 2025, arXiv:2408.11537 [cs]. [Online]. Available: \url{http://arxiv.org/abs/2408.11537}
\BIBentrySTDinterwordspacing

\bibitem{zhao_survey_2025}
\BIBentryALTinterwordspacing
W.~X. Zhao, K.~Zhou, J.~Li, T.~Tang, X.~Wang, Y.~Hou, Y.~Min, B.~Zhang, J.~Zhang, Z.~Dong, Y.~Du, C.~Yang, Y.~Chen, Z.~Chen, J.~Jiang, R.~Ren, Y.~Li, X.~Tang, Z.~Liu, P.~Liu, J.-Y. Nie, and J.-R. Wen, ``A {Survey} of {Large} {Language} {Models},'' Mar. 2025, arXiv:2303.18223 [cs]. [Online]. Available: \url{http://arxiv.org/abs/2303.18223}
\BIBentrySTDinterwordspacing

\bibitem{wang2024llm}
P.~Wang, M.~Robbiani, and Z.~Guo, ``Llm granularity for on-the-fly robot control,'' \emph{arXiv preprint arXiv:2406.14653}, 2024.

\bibitem{xiao2023llm}
H.~Xiao and P.~Wang, ``Llm a*: Human in the loop large language models enabled a* search for robotics,'' \emph{arXiv preprint arXiv:2312.01797}, 2023.

\bibitem{ho_denoising_2020}
\BIBentryALTinterwordspacing
J.~Ho, A.~Jain, and P.~Abbeel, ``Denoising {Diffusion} {Probabilistic} {Models},'' Dec. 2020, arXiv:2006.11239 [cs]. [Online]. Available: \url{http://arxiv.org/abs/2006.11239}
\BIBentrySTDinterwordspacing

\bibitem{wang2024robot}
P.~Wang, Z.~Guo, A.~L. Sait, and M.~H. Pham, ``Robot shape and location retention in video generation using diffusion models,'' in \emph{2024 IEEE/RSJ International Conference on Intelligent Robots and Systems (IROS)}.\hskip 1em plus 0.5em minus 0.4em\relax IEEE, 2024, pp. 7375--7382.

\bibitem{kerbl_3d_2023}
\BIBentryALTinterwordspacing
B.~Kerbl, G.~Kopanas, T.~Leimkühler, and G.~Drettakis, ``{3D} {Gaussian} {Splatting} for {Real}-{Time} {Radiance} {Field} {Rendering},'' Aug. 2023, arXiv:2308.04079 [cs]. [Online]. Available: \url{http://arxiv.org/abs/2308.04079}
\BIBentrySTDinterwordspacing

\bibitem{mildenhall_nerf_2020}
\BIBentryALTinterwordspacing
B.~Mildenhall, P.~P. Srinivasan, M.~Tancik, J.~T. Barron, R.~Ramamoorthi, and R.~Ng, ``{NeRF}: {Representing} {Scenes} as {Neural} {Radiance} {Fields} for {View} {Synthesis},'' Aug. 2020, arXiv:2003.08934 [cs]. [Online]. Available: \url{http://arxiv.org/abs/2003.08934}
\BIBentrySTDinterwordspacing

\bibitem{chiDiffusionPolicyVisuomotor2024}
\BIBentryALTinterwordspacing
C.~Chi, Z.~Xu, S.~Feng, E.~Cousineau, Y.~Du, B.~Burchfiel, R.~Tedrake, and S.~Song, ``Diffusion {Policy}: {Visuomotor} {Policy} {Learning} via {Action} {Diffusion},'' Mar. 2024, arXiv:2303.04137 [cs]. [Online]. Available: \url{http://arxiv.org/abs/2303.04137}
\BIBentrySTDinterwordspacing

\bibitem{he_deep_2015}
\BIBentryALTinterwordspacing
K.~He, X.~Zhang, S.~Ren, and J.~Sun, ``Deep {Residual} {Learning} for {Image} {Recognition},'' Dec. 2015, arXiv:1512.03385 [cs]. [Online]. Available: \url{http://arxiv.org/abs/1512.03385}
\BIBentrySTDinterwordspacing

\bibitem{chi_universal_2024}
\BIBentryALTinterwordspacing
C.~Chi, Z.~Xu, C.~Pan, E.~Cousineau, B.~Burchfiel, S.~Feng, R.~Tedrake, and S.~Song, ``\BIBforeignlanguage{en}{Universal {Manipulation} {Interface}: {In}-{The}-{Wild} {Robot} {Teaching} {Without} {In}-{The}-{Wild} {Robots}},'' in \emph{\BIBforeignlanguage{en}{Robotics: {Science} and {Systems} {XX}}}.\hskip 1em plus 0.5em minus 0.4em\relax Robotics: Science and Systems Foundation, Jul. 2024. [Online]. Available: \url{http://www.roboticsproceedings.org/rss20/p045.pdf}
\BIBentrySTDinterwordspacing

\bibitem{krizhevsky_imagenet_2017}
\BIBentryALTinterwordspacing
A.~Krizhevsky, I.~Sutskever, and G.~E. Hinton, ``\BIBforeignlanguage{en}{{ImageNet} classification with deep convolutional neural networks},'' \emph{\BIBforeignlanguage{en}{Communications of the ACM}}, vol.~60, no.~6, pp. 84--90, May 2017. [Online]. Available: \url{https://dl.acm.org/doi/10.1145/3065386}
\BIBentrySTDinterwordspacing

\bibitem{ridnik_imagenet-21k_2021}
\BIBentryALTinterwordspacing
T.~Ridnik, E.~Ben-Baruch, A.~Noy, and L.~Zelnik-Manor, ``{ImageNet}-{21K} {Pretraining} for the {Masses},'' Aug. 2021, arXiv:2104.10972 [cs]. [Online]. Available: \url{http://arxiv.org/abs/2104.10972}
\BIBentrySTDinterwordspacing

\bibitem{kim_openvla_2024}
\BIBentryALTinterwordspacing
M.~J. Kim, K.~Pertsch, S.~Karamcheti, T.~Xiao, A.~Balakrishna, S.~Nair, R.~Rafailov, E.~Foster, G.~Lam, P.~Sanketi, Q.~Vuong, T.~Kollar, B.~Burchfiel, R.~Tedrake, D.~Sadigh, S.~Levine, P.~Liang, and C.~Finn, ``{OpenVLA}: {An} {Open}-{Source} {Vision}-{Language}-{Action} {Model},'' Sep. 2024, arXiv:2406.09246 [cs]. [Online]. Available: \url{http://arxiv.org/abs/2406.09246}
\BIBentrySTDinterwordspacing

\bibitem{simeoni2025dinov3}
O.~Sim{\'e}oni, H.~V. Vo, M.~Seitzer, F.~Baldassarre, M.~Oquab, C.~Jose, V.~Khalidov, M.~Szafraniec, S.~Yi, M.~Ramamonjisoa \emph{et~al.}, ``Dinov3,'' \emph{arXiv preprint arXiv:2508.10104}, 2025.

\bibitem{florence_implicit_2021}
\BIBentryALTinterwordspacing
P.~Florence, C.~Lynch, A.~Zeng, O.~Ramirez, A.~Wahid, L.~Downs, A.~Wong, J.~Lee, I.~Mordatch, and J.~Tompson, ``Implicit {Behavioral} {Cloning},'' Sep. 2021, arXiv:2109.00137 [cs]. [Online]. Available: \url{http://arxiv.org/abs/2109.00137}
\BIBentrySTDinterwordspacing

\bibitem{mandlekar_what_2021}
\BIBentryALTinterwordspacing
A.~Mandlekar, D.~Xu, J.~Wong, S.~Nasiriany, C.~Wang, R.~Kulkarni, L.~Fei-Fei, S.~Savarese, Y.~Zhu, and R.~Martín-Martín, ``What {Matters} in {Learning} from {Offline} {Human} {Demonstrations} for {Robot} {Manipulation},'' Sep. 2021, arXiv:2108.03298 [cs]. [Online]. Available: \url{http://arxiv.org/abs/2108.03298}
\BIBentrySTDinterwordspacing

\bibitem{ronneberger_u-net_2015}
\BIBentryALTinterwordspacing
O.~Ronneberger, P.~Fischer, and T.~Brox, ``U-{Net}: {Convolutional} {Networks} for {Biomedical} {Image} {Segmentation},'' May 2015, arXiv:1505.04597 [cs]. [Online]. Available: \url{http://arxiv.org/abs/1505.04597}
\BIBentrySTDinterwordspacing

\bibitem{perez_film_2018}
\BIBentryALTinterwordspacing
E.~Perez, F.~Strub, H.~d. Vries, V.~Dumoulin, and A.~Courville, ``\BIBforeignlanguage{en}{{FiLM}: {Visual} {Reasoning} with a {General} {Conditioning} {Layer}},'' \emph{\BIBforeignlanguage{en}{Proceedings of the AAAI Conference on Artificial Intelligence}}, vol.~32, no.~1, Apr. 2018, number: 1. [Online]. Available: \url{https://ojs.aaai.org/index.php/AAAI/article/view/11671}
\BIBentrySTDinterwordspacing

\end{thebibliography}

\end{document}